\documentclass[11pt,draftcls,onecolumn]{IEEEtran}
\usepackage{subfig}
\usepackage{amssymb}
\usepackage{cite}
\usepackage[pdftex]{graphicx}
\graphicspath{{./}{figures/}}
\usepackage[table]{xcolor}
\usepackage{arydshln}
\usepackage{amsmath}
\usepackage{amssymb}
\usepackage{url}
\usepackage{setspace}
\usepackage{fixltx2e}
\usepackage{rotating}
\usepackage{multirow}
\usepackage{floatrow}
\usepackage[font=small,labelfont=bf]{caption}
\newfloatcommand{capbtabbox}{table}[][\FBwidth]

\newcolumntype{M}{>{$\vcenter\bgroup\hbox\bgroup}c<{\egroup\egroup$}}

\usepackage{blindtext}

\def\ie{\emph{i.e. }}

\newcommand{\im}[1]{\mathbf{#1}}
\newcommand{\est}[1]{\widehat{#1}}

\usepackage{algorithm}
\usepackage{algorithmic}
\usepackage{caption}
\begin{document}
\title{James-Stein Type Center Pixel Weights for Non-Local Means Image Denoising}
\author{Yue~Wu,~\IEEEmembership{Member}, {Brian~Tracey},~\IEEEmembership{Senior Member,}
        and Joseph~P.~Noonan,~\IEEEmembership{Life Member,~IEEE,}
\thanks{ Yue Wu, Brian~Tracey and Joseph P. Noonan are with the Department of Electrical and Computer Engineering, Tufts University, Medford, MA 02155, United States; e-mail: ywu03@ece.tufts.edu.}}
\maketitle

\begin{abstract}
Non-Local Means (NLM) and variants have been proven to be effective and robust in many image denoising tasks. In this letter, we study the parameter selection problem of center pixel weights (CPW) in NLM. Our key contributions are: 1) we give a novel formulation of the CPW problem from the statistical shrinkage perspective; 2) we introduce the James-Stein type CPWs for NLM; and 3) we propose a new adaptive CPW that is locally tuned for each image pixel. Our experimental results showed that compared to existing CPW solutions, the new proposed CPWs are more robust and effective under various noise levels. In particular, the NLM with the James-Stein type CPWs attain higher means with smaller variances in terms of the peak signal and noise ratio, implying they improve the NLM robustness and make it less sensitive to parameter selection.
\end{abstract}

\IEEEpeerreviewmaketitle

\section{Introduction}
Image noise commonly exists in the image acquisition, quantization, transmission and many other processing stages. A digital image contaminated by noises leads to visible loss in image quality and can impact many advanced image processing and computer vision tasks like tracking, recognition, classification, etc. The importance of image denoising is therefore self-explanatory.

Conventional image denoising methods more or less related to filters \cite{NLM0}, like moving average filters, Wiener filters, and wavelet filter banks. These filter-based image denoising techniques are commonly of low complexity and can be easily implemented in hardware. However, their performance is not always adequate. With the help of the increased computational capacity powered by digital processors, many advanced denoising techniques are now feasible. Among these techniques, the Non-Local Means (NLM) method \cite{NLM0,NLM1} has attracted significant attention in recent years. The NLM denoises an image pixel as the weighted sum of its noisy neighbors, where each weight reflects the similarity between the local patch centered at the noisy pixel to be denoised and the patch centered at the neighbor pixel. In this way, NLM adapts the denoising process for each pixel and thus outperforms conventional techniques \cite{NLM0}.

Many improvements on the original NLM have been proposed in recent years. These discussions mainly focus on three questions: 1) how to pick NLM parameters heuristically or automatically \cite{SURE,CPW}; 2) how to accelerate the NLM or save computations without loss of denoising performance \cite{IINLM,5710056,6324435}; and 3) how to adjust the NLM framework to achieve better performance \cite{5356177,6036056,6315713,5548921,multipatch}. In \cite{SURE}, the closed-form of Stein's risk estimator is derived for the NLM and allows prediction of the risk of a denoised image without knowing the clean image. In \cite{IINLM}, the integral image scheme is adapted to the NLM and greatly reduces the computational costs. In \cite{multipatch}, multiple-patches are discussed to eliminate artifacts in the NLM.

The importance of the center pixel weight (CPW) has been noticed for a long time \cite{NLM0,CPW}, and various heuristic weights are designed and used \cite{CPW}. However, the methods proposed are non-ideal as they do not consider all aspects of CPW problem (see details in Section III-A). Thus, new CPWs need to be designed. In this letter, we discuss the CPW problem in NLM and propose new solutions based on the James-Stein estimator \cite{JamesStein}. The rest of the letter is organized as: Sec. II reviews the NLM and related works on CPWs; Sec. III discusses the new formulation of the CPW problem and the new James-Stein type CPWs; Sec. IV shows the experimental comparisons; and we conclude the letter in Sec. V.

\section{A Brief Review}
\subsection{The Classic Non-Local Mean Algorithm}
Let $\im{x}\!=\!\{\!x_l\!\}_{l\in\mathbb I}$ be a two dimensional clean image defined on the spatial domain $\mathbb{I}$, with $l\!=\!(l_1,l_2)$ the $l$th pixel located at the intersection of the $l_1$th row and the $l_2$th column. Given one of $\im{x}$'s noisy observation $\im{y}\!=\!\{\!y_l\!\}_{l\in\mathbb I}$,
whose pixels are contaminated by an i.i.d. zero-mean Gaussian noise with a variance of $\sigma^2$, namely
\begin{equation}\label{eq.gaussiannoise}
    y_l = x_l+n_l, \textrm{\, and \,} n_l\sim{\cal{N}}(0,\sigma^2).
\end{equation}
The classic NLM method \cite{NLM0,NLM1} estimates the clean pixel $x_l$ by using all pixels within a prescribed search region $\mathbb{S}$, typically a square or a rectangular region. Specifically, the estimated $\widehat{x_l}$ is the weighted sum of $y_l$'s noisy neighbors as
\begin{equation}\label{eq.nlm}
\est{x_l} = \sum_{k\in\mathbb{S}}\dfrac{w_{l,k}y_{k}}{\sum_{k\in\mathbb{S}}w_{l,k}}
\end{equation}
where each weight is computed by quantifying the similarity between two local patches around noisy pixels $y_l$ and $y_k$ as shown in Eq. \eqref{eq.nlmweight}, where $G_\alpha$ is a Gaussian weakly smooth kernel \cite{NLM0}, $\mathbb{P}$ denotes the local patch, typically a square centered at the pixel (for example a $\mathbb{P}\!=\![-1,0,1]\!\times\![-1,0,1]$ denotes a 3$\!\times\!$3 patch).  The parameter $h$ can be considered as a temperature parameter controlling the behavior of the weight function, namely as $h\!\rightarrow\!0$, all weights approach to 1; while as $h\!\rightarrow\!\infty$, all weights approach to zero.
\begin{equation}\label{eq.nlmweight}
    w_{l,k} = \exp\left(-{\textstyle\sum_{j\in\mathbb{P}} G_\alpha({y_{l+j}-y_{k+j})^2}/ h}\right)
\end{equation}
\subsection{Existing Central Pixel Weights}
The CPW in the classic NLM is unitary, because \eqref{eq.nlmweight} implies $w_{l,l}\!=\!1$ for all $l\!\in\!\mathbb{I}$. However, this unitary CPW is reported not to perform well in many cases \cite{CPW}. Indeed, in the case that $y_l$ is really noisy and $y_l$ has a rare patch (implying that for all $k\in\mathbb{S}\backslash l$ non-center weights $w_{l,k}$ are very small compared to 1), a unitary CPW means the contribution of the noisy pixel $y_l$ dominates in the denoised pixel $\est{x_l}$, implying a poor performance.

In addition to this CPW, several other CPWs have been proposed and used in the NLM community to enhance performance.  These include the zero CPW (Eq. \eqref{eq.cpw0}), the Stein CPW (Eq. \eqref{eq.cpwstein}), the max CPW (Eq. \eqref{eq.cpwmax}) and the heuristic CPW (Eq. \eqref{eq.cpwheur}). In the rest of the letter, we use $v_l$ to denote these existing CPWs.
\begin{equation}\label{eq.cpw1}
    v_l^{one} = 1
\end{equation}
\begin{equation}\label{eq.cpw0}
    v_l^{zero} = 0
\end{equation}
\begin{equation}\label{eq.cpwstein}
    v_l^{stein} = \exp(-\sigma^2|\mathbb{P}|/h)
\end{equation}
\begin{equation}\label{eq.cpwmax}
    v_l^{max} = \max_{k\in\mathbb{S}\backslash l} w_{l,k}
\end{equation}
\begin{equation}\label{eq.cpwheur}
    v_l^{heru.} = \left\{\begin{array}{rl}\infty,&\textrm{\, if \,} v_l^{max}\leq threshold\\
    v_l^{max},&\textrm{otherwise}\end{array}\right.
\end{equation}

These CPWs can be classified into two groups: global CPWs (Eqs. \eqref{eq.cpw1},\eqref{eq.cpw0} and \eqref{eq.cpwstein}) and local CPWs (Eqs. \eqref{eq.cpwmax} and \eqref{eq.cpwheur}). The global CPWs use a constant CPW for all pixels, while the local CPWs use different CPWs for pixels. In the next section, we will show that these CPWs fail to take all variables into full consideration and therefore oversimplify the CPW problem.

\section{Shrinkage based Center Pixel Weights}
\subsection{The CPW problem in the form of shrinkage estimator}
To fully reveal the CPW problem, we separate the contributions of the non-center and of the center pixels in the NLM denoised pixel $\est{x_l}$ (Eq. \eqref{eq.nlm})
\begin{eqnarray}\label{eq.estxl2}
    \est{x_l} = \dfrac{W_l}{W_l+v_l}\est{z_l}+\dfrac{v_l}{W_l+v_l}y_l
\end{eqnarray}
where $W_l$ is the summation of all non-center weights
\begin{equation}\label{eq.Wl}
    \textstyle W_l =\sum_{k\in\mathbb{S}\backslash l}w_{l,k}
\end{equation}
and $\est{z_l}$ is the denoised pixel by using all non-center weights.
\begin{equation}\label{eq.1st}
    \textstyle \est{z_l} = \sum_{k\in\mathbb{S}\backslash l}{w_{l,k}y_{k}}/{W_l}.
\end{equation}

As one can see, a CPW $v_l$ is contained in the coefficients of both $\est{z_l}$ and $y_l$ and thus influences the final denoised pixel $\est{x_l}$ directly. If we are given an optimal $\widehat{x_l}$ and solve for $v_l$, then it is clear that the optimal $v_l$ should be a function of $W_l$, $\est{z_l}$ and $y_l$. In other words, a CPW $v_l$ that does not consider  all these variables is incomplete. It is noticeable that the global CPWs $v_l^{one}$ $v_l^{zero}$ and $v_l^{stein}$ neglects all three, while the local CPWs $v_l^{max}$ and $v_l^{heur.}$ neglect $y_l$.

Let $p_l$ be the fraction ($p_l\in[0,1]$) of the contribution of the center pixel $y_l$ in $\widehat{x_l}$, namely
\begin{equation}\label{eq.pl}
    p_l = v_l/(v_l+W_l).
\end{equation}
$p_l$ is then a normalized version of $v_l$. Consequently, the NLM-CPW problem in \eqref{eq.estxl2} can be rewritten as
\begin{equation}\label{eq.estxl3}
    \est{x_l} = (1-p_l)\est{z_l}+p_ly_l
\end{equation}
and is a so-called shrinkage estimator, which improves an existing estimator by using the raw data. In the context of the NLM, the existing estimator is $\est{z_l}$ and the raw data is the noisy pixel $y_l$. The effect of the CPW is to tune the final denoised pixel $\est{x_l}$ in somewhere between $\est{z_l}$ and $y_l$, or equivalently to shrink $y_l$ towards to $\est{z_l}$.

\subsection{The James-Stein Center Pixel Weight}
One important result in shrinkage estimators is the James-Stein estimator \cite{JamesStein}. It states that for an unknown parameter vector $\im{a}$ and its observations of $\im{b}$ with the relation,
\begin{equation}\label{eq.js}
    \im{b}\sim{{\cal N}(\im{a},\sigma^2\im{I})}
\end{equation}
there exists a James-Stein estimator that shrinks towards an arbitrary vector $\im{c}$ in the form that
\begin{equation}\label{eq.jse}
    \im{\est{a}}^{JS} = \im{c}+ q(\im{b}-\im{c})= (1-q)\im{c}+q\im{b}
\end{equation}
with the coefficient $q$ of form \eqref{eq.jsvl} \cite{JamesStein}.
\begin{equation}\label{eq.jsvl}
    q = 1-{(m-2)\sigma^2}/{\|\im{b}-\im{{c}}\|^2}
\end{equation}
This James-Stein estimator is a classic solution to minimize the risk of estimation in terms of the mean square error  $E[\|\im{a}-\im{\est{a}}\|^2]$ \cite{Stein}, where $\|.\|$ denotes the $L^2$-norm.

In the context of NLM-CPW problem, the James-Stein based CPW (JSCPW) has the weight of form \eqref{eq.jsvl},
\begin{equation}\label{eq.jsvl}
    p^{*JS} = 1-{(m-2)\sigma^2}/{\|\im{y}-\im{\est{z}}\|^2}
\end{equation}
where $m\!\!=\!\!|\mathbb{I}|$ is the number of pixels in the image, and the corresponding new estimator is
\begin{equation}\label{eq.estxl3}
    \im{\est{x}}^{JS} = (1-p^{*JS})\im{\est{z}}+p^{*JS}\im{y}.
\end{equation}
\subsection{Local Adapted James-Stein Center Pixel Weights}
Although the proposed JSCPW considers all $W_l$, $\est{x_l}$ and $y_l$, it is still a global CPW, which gives the weight to each pixel unbiasedly. However, we know the denoising process is always biased rather than unbiased for each pixel. For example, pixels in a homogeneous region are commonly better denoised than edge pixels. Therefore, ideally we want a locally adapted CPW for each pixel as $p_l$ in the form of \eqref{eq.estxl3}. One natural idea is to replace ${\|\im{y}-\im{\est{z}}\|^2}$ in \eqref{eq.estxl3} with ${\|y_l-{\est{z_l}}\|^2}$, but it does not lead to a stable solution, because ${\|y_l-{\est{z_l}}\|^2}$ is commonly noisy. Alternatively, we view each image block as a small image and thus the JSCPW \eqref{eq.jsvl} computed for a local block gives a local CPW adapted to each pixel.

Without loss of generality, say $\est{\im{zb}}_l\!=\!\{\est{z_{l+b}}| \forall b\in\mathbb{B}\}$ and ${\im{yb}}_l\!=\!\{{y_{l+b}}| \forall b\in\mathbb{B}\}$, so there are two corresponding local image blocks around the $l$th pixel in $\im{\est{z}}$ and $\im{y}$, respectively. Given a prescribed local block region $\mathbb{B}$, then the local James-Stein center pixel weight (LJSCPW) can be computed as
\begin{equation}\label{}
    p_l^{*LJS} = 1-{(|\mathbb{B}|-2)\sigma^2}/{\|\im{yb}_l-\im{\est{z b}}_l\|^2}.
\end{equation}
In this way, we construct a local CPW for each pixel, and thus the denoised pixel by using LJSCPW can be written as
\begin{equation}\label{eq.estxl3}
    \est{x_l}^{LJS} = (1-p_l^{*LJS})\est{z_l}+p_l^{*LJS}y_l.
\end{equation}
Intuitively, this LJSCPW helps eliminate the influence of remote image pixels and tunes the optimization locally.

\subsection{Implementation}
The computational cost of the LJSCPW can be only 5 operations/pixel more than the existing CPWs. Specifically, we construct the integral image \cite{IINLM} $\im{II}$ with 2 operations/pixel for the pixel-wise mean square error between $\im{\est{z}}$ and $\im{y}$. Each pixel is the summation of form \eqref{eq.ii}.
\begin{equation}\label{eq.ii}
  \textstyle  II_{l} = \sum_{i=\{(i_1,i_2)|i_1\in[1,l_1],i_2\in[1,l_2]\}}(y_{i}-\est{z_{i}})^2,
\end{equation}
This integral image $\im{II}$ then allows computation of $\|\!\im{yb}_l\!-\!\im{\est{z b}}_l\!\|^2$ for an arbitrary rectangular $\mathbb{B}$ with 3 operations/pixel.

In a real case, one may see $p_l^{*LJS}\!<\!0$ which conflicts with our assumption that $p_l^{*LJS}\in [0,1]$. This case happens when $\im{\est{z r}}_l$ is a slightly denoised version of $\im{yr}_l$. So it is reasonable to use $\est{z_l}$ rather than $y_l$, implying $p_l^{*LJS}=0$. Therefore, we use the positive part of $p_l^{*LJS}$ in Eq. \eqref{eq.plljsp}, where $(.)^+ = \max(.,0)$.
 \begin{equation}\label{eq.plljsp}
    p_l^{*LJS+} = \left(1-{(|\mathbb{B}|-2)\sigma^2}/{\|\im{yb}_l-\im{\est{z b}}_l\|^2}\right)^{+}
\end{equation}
Thus $p_l^{*LJS+}$ ranges from $[0,1]$.  Because $p_l^{*LJS+}=1$ would indicate the raw data is used, i.e. the pixel is not denoised, it may prove useful in some applications to limit $p_l^{*LJS+}$ to a user-defined value less than unity.  In this letter, however, we allow the shrinkage operator to operate over the full range.

\section{Simulation Results}
All following simulations are done under the MATLAB r2010a environment. We compare the performance of CPWs \ie $v_l^{one},v_l^{zero},v_l^{stein},v_l^{max},v_l^{heur.}$ with the proposed James-Stein type CPWs $p^{*JS}$ and $p_l^{*LJS}$ under the classic NLM framework (only CPW part is changed). In particular, we set searching region $\mathbb{S}$ to 31$\times$31 square and use $\mathbb{B}\!=\!\mathbb{P}$ for $p_l^{*LJS}$ in the experiment. All test images \footnotemark[1] are gray-scale images with additive Gaussian noises of $\sigma\!\! \!\in\!\{\!10,20,40\!\}$. For each test image, we denoise it by using the 200 temperature parameters $h$ ranging from 1\% to 200\% of $\sigma^2{|\mathbb{P}|}$. The denoising performance for each CPW scheme is then evaluated by computing its mean and standard deviation in terms of the peak signal noise ratio (PSNR) (dB) defined in Eq. \eqref{eq.psnr}.
\begin{equation}\label{eq.psnr}
    \textrm{PSNR}(\im{x},\im{\est{x}}) = 20\log_{10}255-10\log_{10}\|\im{x}-\im{\est{x}}\|^2/|\mathbb{I}|
\end{equation}

\footnotetext[1]{are available under the page \url{http://www.cs.tut.fi/~foi/GCF-BM3D/BM3D_images.zip} as the date of 10/09/2012.}

These results are summarized in Table \ref{tab.comp}. As one can see, compared to the PSNR performance of the zero CPW, the proposed $p_l^{*LJS}$ (LJSCPW) and $p^{*JS}$ (JSCPW) are the only two in all CPWs that always improve the denoising performance in terms of a higher mean PSNR with a smaller variance, regardless of patch sizes, test images and noise levels. This implies that James-Stein type CPWs are more efficient than other CPWs.

\begin{table}[!h]
\caption{(mean$\pm$ standard deviation) PSNR (dB) comparisons for various center pixel weighting schemes}\label{tab.comp}
\tiny\centering
\begin{tabular}{@{}c@{}c@{}m{.025cm}@{}||@{}m{.025cm}@{}r@{$\pm$}l@{}m{.025cm}@{}|@{}m{.025cm}@{}r@{$\pm$}l@{}m{.025cm}@{}|@{}m{.025cm}@{}r@{$\pm$}l@{}m{.025cm}@{}|@{}m{.025cm}@{}r@{$\pm$}l@{}m{.025cm}@{}|@{}m{.025cm}@{}r@{$\pm$}l@{}m{.025cm}@{}|@{}m{.025cm}@{}r@{$\pm$}l@{}m{.025cm}@{}|@{}m{.025cm}@{}r@{$\pm$}l@{}m{.025cm}@{}||@{}m{.025cm}@{}r@{$\pm$}l@{}m{.025cm}@{}|@{}m{.025cm}@{}r@{$\pm$}l@{}m{.025cm}@{}|@{}m{.025cm}@{}r@{$\pm$}l@{}m{.025cm}@{}|@{}m{.025cm}@{}r@{$\pm$}l@{}m{.025cm}@{}|@{}m{.025cm}@{}r@{$\pm$}l@{}m{.025cm}@{}|@{}m{.025cm}@{}r@{$\pm$}l@{}m{.025cm}@{}|@{}m{.025cm}@{}r@{$\pm$}l@{}m{.025cm}@{}}
\hline
&&&\multicolumn{28}{c||}{\bf 5$\times$5 Patch $ \mathbb{P}$}&\multicolumn{28}{c}{\bf 7$\times$7 Patch $ \mathbb{P}$}\\
&&&\multicolumn{4}{c|}{$v_l^{zero}$}&\multicolumn{4}{c|}{$v_l^{one}$}&\multicolumn{4}{c|}{$v_l^{stein}$}&\multicolumn{4}{c|}{$v_l^{max}$}&\multicolumn{4}{c|}{$v_l^{heur.}$}&\multicolumn{4}{c|}{$p^{*JS}$}&\multicolumn{4}{c||}{$p_l^{*LJS}$}&\multicolumn{4}{c|}{$v_l^{zero}$}&\multicolumn{4}{c|}{$v_l^{one}$}&\multicolumn{4}{c|}{$v_l^{stein}$}&\multicolumn{4}{c|}{$v_l^{max}$}&\multicolumn{4}{c|}{$v_l^{heur.}$}&\multicolumn{4}{c|}{$p^{*JS}$}&\multicolumn{4}{c}{$p_l^{*LJS}$}\\\hline
&$\sigma$&&\multicolumn{56}{c}{\bf 256$\times$256 cameraman}\\\hline
&\bf{10}&&&28.14&2.73&&&31.72&1.07&&&31.69&0.80&&&30.92&0.75&&&31.22&1.03&&&30.70&0.97&&&32.75&0.56&&&26.18&3.41&&&31.27&1.08&&&31.38&0.80&&&30.03&0.65&&&30.61&1.12&&&30.26&0.90&&&32.59&0.50&\\
&\bf{20}&&&27.25&1.28&&&27.73&1.61&&&27.93&1.05&&&27.74&1.02&&&27.75&1.05&&&27.66&1.08&&&28.50&0.71&&&26.31&1.62&&&27.30&1.67&&&27.69&1.13&&&27.28&1.09&&&27.28&1.23&&&27.26&1.05&&&28.57&0.66&\\
&\bf{40}&&&23.76&1.17&&&23.35&2.03&&&23.79&1.18&&&23.75&1.27&&&23.74&1.29&&&23.90&1.08&&&24.10&0.96&&&23.33&1.28&&&22.77&2.20&&&23.44&1.31&&&23.43&1.32&&&23.38&1.46&&&23.69&1.07&&&24.31&0.87&\\
\hline
&$\sigma$&&\multicolumn{56}{c}{\bf 256$\times$256 house}\\\hline
&\bf{10}&&&33.30&2.03&&&33.68&1.58&&&34.08&0.94&&&33.97&0.94&&&33.97&0.99&&&33.77&1.36&&&34.34&0.73&&&32.73&2.68&&&33.34&1.70&&&34.01&0.88&&&33.86&0.89&&&33.82&1.09&&&33.69&1.36&&&34.45&0.69&\\
&\bf{20}&&&30.07&1.20&&&29.60&2.10&&&30.11&1.18&&&30.06&1.27&&&30.05&1.30&&&30.16&1.14&&&30.09&0.98&&&29.87&1.49&&&29.18&2.34&&&30.03&1.38&&&30.01&1.38&&&29.95&1.55&&&30.15&1.33&&&30.47&1.00&\\
&\bf{40}&&&25.50&1.41&&&25.10&2.47&&&25.67&1.40&&&25.57&1.56&&&25.57&1.57&&&25.70&1.38&&&25.68&1.17&&&25.78&1.55&&&24.79&2.78&&&25.79&1.55&&&25.72&1.59&&&25.68&1.75&&&25.92&1.44&&&25.97&1.24&\\
\hline
&$\sigma$&&\multicolumn{56}{c}{\bf 256$\times$256 peppers}\\\hline
&\bf{10}&&&30.41&2.21&&&31.98&1.26&&&32.13&0.91&&&31.80&0.86&&&31.87&0.94&&&31.65&1.00&&&32.75&0.60&&&29.15&3.11&&&31.60&1.30&&&32.03&0.93&&&31.37&0.88&&&31.48&1.12&&&31.55&1.01&&&32.61&0.62&\\
&\bf{20}&&&27.74&1.12&&&27.54&1.68&&&27.92&1.09&&&27.89&1.10&&&27.88&1.13&&&28.04&0.95&&&28.38&0.80&&&27.16&1.49&&&27.06&1.83&&&27.64&1.31&&&27.59&1.27&&&27.53&1.41&&&27.86&1.02&&&28.41&0.84&\\
&\bf{40}&&&23.53&1.25&&&23.04&2.01&&&23.55&1.25&&&23.48&1.32&&&23.48&1.34&&&23.68&1.13&&&23.67&0.98&&&23.38&1.46&&&22.60&2.26&&&23.41&1.46&&&23.39&1.46&&&23.34&1.61&&&23.73&1.20&&&23.89&1.08&\\
\hline
&$\sigma$&&\multicolumn{56}{c}{\bf 512$\times$512 lenna}\\\hline
&\bf{10}&&&32.75&1.58&&&32.90&1.45&&&33.18&0.94&&&33.19&0.94&&&33.18&0.97&&&33.22&1.06&&&33.65&0.67&&&32.02&2.36&&&32.57&1.57&&&33.09&1.02&&&33.01&1.03&&&32.95&1.18&&&33.13&1.13&&&33.83&0.67&\\
&\bf{20}&&&29.64&1.05&&&29.16&1.96&&&29.66&1.04&&&29.61&1.14&&&29.60&1.16&&&29.73&0.99&&&29.67&0.89&&&29.40&1.20&&&28.73&2.17&&&29.50&1.14&&&29.50&1.15&&&29.44&1.34&&&29.71&1.05&&&29.97&0.86&\\
&\bf{40}&&&25.96&1.31&&&25.40&2.48&&&25.96&1.31&&&25.86&1.50&&&25.86&1.51&&&25.96&1.31&&&25.96&1.13&&&26.21&1.25&&&25.25&2.86&&&26.22&1.25&&&26.15&1.36&&&26.11&1.55&&&26.30&1.19&&&26.23&1.11&\\
\hline
&$\sigma$&&\multicolumn{56}{c}{\bf 512$\times$512 barbara}\\\hline
&\bf{10}&&&31.44&1.83&&&31.84&1.28&&&32.04&0.96&&&32.10&0.93&&&32.09&0.97&&&32.15&0.92&&&32.69&0.62&&&30.62&2.83&&&31.62&1.37&&&32.07&1.02&&&31.97&0.98&&&31.91&1.11&&&32.24&1.01&&&32.90&0.65&\\
&\bf{20}&&&27.74&1.17&&&27.29&1.66&&&27.77&1.16&&&27.75&1.17&&&27.74&1.20&&&28.02&0.95&&&28.13&0.85&&&27.70&1.44&&&27.08&1.88&&&27.82&1.39&&&27.86&1.39&&&27.79&1.52&&&28.27&1.10&&&28.56&0.93&\\
&\bf{40}&&&23.72&1.09&&&23.21&1.96&&&23.72&1.09&&&23.64&1.19&&&23.73&1.20&&&23.82&1.01&&&23.77&0.93&&&23.92&1.28&&&23.05&2.30&&&23.93&1.28&&&23.88&1.29&&&23.83&1.47&&&24.20&1.10&&&24.15&1.03&\\
\hline
&$\sigma$&&\multicolumn{56}{c}{\bf 512$\times$512 boat}\\\hline
&\bf{10}&&&30.30&1.98&&&31.15&1.12&&&31.25&0.96&&&31.21&0.97&&&31.21&1.00&&&31.50&0.87&&&32.14&0.51&&&29.22&2.75&&&30.68&1.08&&&30.90&1.02&&&30.73&1.07&&&30.70&1.15&&&31.47&0.86&&&32.08&0.52&\\
&\bf{20}&&&27.54&1.06&&&27.26&1.57&&&27.61&1.05&&&27.61&1.07&&&27.60&1.10&&&27.90&0.84&&&28.12&0.73&&&26.89&1.34&&&26.64&1.64&&&27.14&1.20&&&27.16&1.22&&&27.10&1.33&&&27.66&0.91&&&28.08&0.71&\\
&\bf{40}&&&24.18&1.05&&&23.70&2.03&&&24.19&1.05&&&24.12&1.18&&&24.11&1.19&&&24.26&1.00&&&24.25&0.91&&&24.03&1.09&&&23.27&2.26&&&24.04&1.09&&&24.01&1.13&&&23.97&1.31&&&24.30&0.94&&&24.38&0.86&\\
\hline
\end{tabular}
\end{table}

We found two general performance trends: 1) as $h\!\!\rightarrow\!\!\infty$, all CPWs except $p_l^{*LJS}$ and $p^{*JS}$ converges together; and 2) as $\sigma$ increases, the impact of CPWs on denoising results decreases. Fig. 1 illustrates these trends in NLM performance for the  \textit{cameraman} image. The first trend occurs because as $h\rightarrow\infty$, all non-center weights go to 1. The CPWs $v_l^{one}$, $v_l^{stein}$, $v_l^{max}$ and $v_l^{heur.}$ are then all 1s  and thus converge. The second trend is because the noisy image $\im{y}$ become less effective in providing a good guide in shrinkage when the noise level is high. Thus the performance differences of various CPWs gets smaller as the noise level increases. However, it is noticeable that the proposed locally adapted $p_l^{*LJS}$ outperforms other CPWs regardless of chosen temperature parameter $h$, and the NLM performance using the proposed CPWs decays much slower than using other CPWs. Similar behaviors are also observed for other patches sizes and test images. Finally, the NLM denoising results for $h=\sigma^2|\mathbb{P}|$ is given in Fig. 2. It is noticeable that the proposed $p_l^{*LJS}$ helps keep weak edges (see the two buildings in the background).
\begin{figure}[!h]
  \scriptsize\centering
  \begin{tabular}{M@{}M@{}M}
     & {\bf 5$\times$5 Patch $ \mathbb{P}$}& {\bf 7$\times$7 Patch $ \mathbb{P}$} \\
    \multirow{1}{*}{\begin{sideways}{\bf $\sigma=$10}\end{sideways}}& \includegraphics[width=6cm]{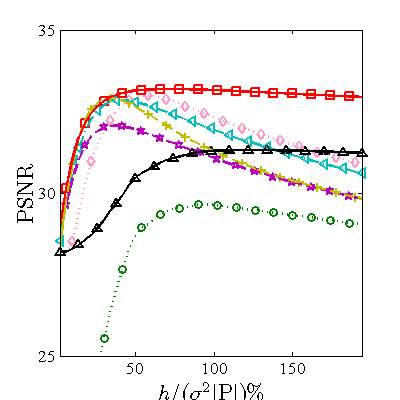} & \includegraphics[width=6cm]{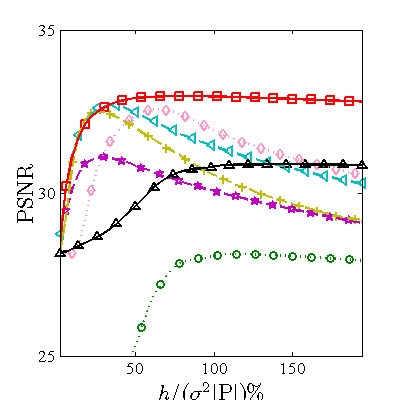} \\
    \multirow{1}{*}{\begin{sideways}{\bf $\sigma=$20}\end{sideways}}  & \includegraphics[width=6cm]{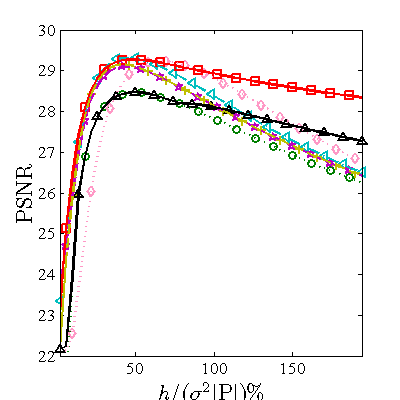} & \includegraphics[width=6cm]{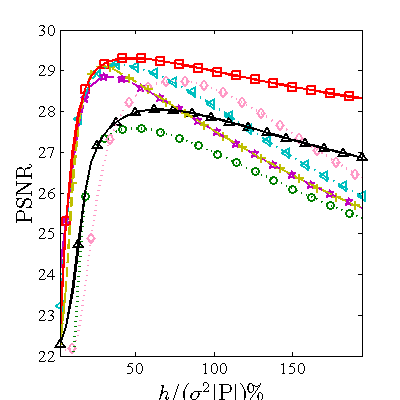} \\
    \multirow{1}{*}{\begin{sideways}{\bf $\sigma=$40}\end{sideways}} & \includegraphics[width=6cm]{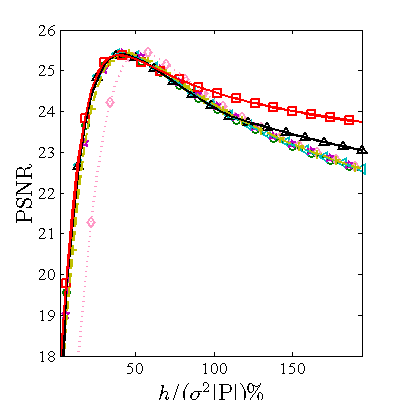}\vspace{10pt} & \includegraphics[width=6cm]{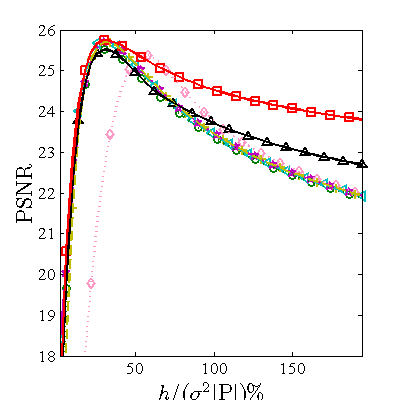} \vspace{10pt}\\
    \multicolumn{3}{c}{\includegraphics[width=12cm]{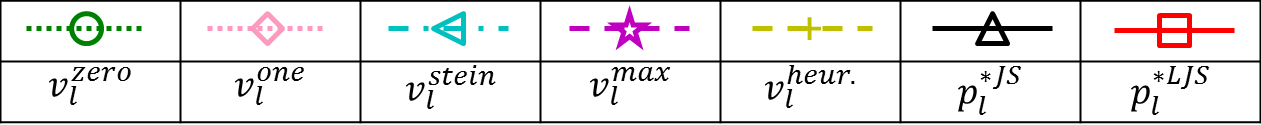}}
  \end{tabular}
  \caption{The PSNR (dB) performance for various CPWs}\vspace{-10pt}
\end{figure}
\begin{figure}[h]
  \scriptsize\centering
  \begin{tabular}{M@{}m{.05cm}@{}M@{}m{.05cm}@{}M}
\includegraphics[width=4cm]{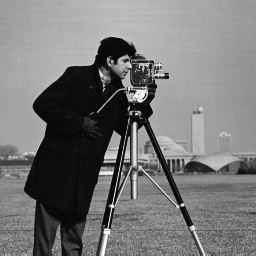} && \includegraphics[width=4cm]{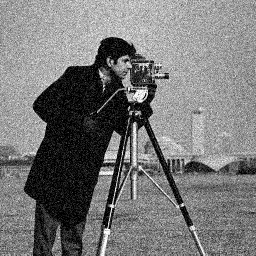} && \includegraphics[width=4cm]{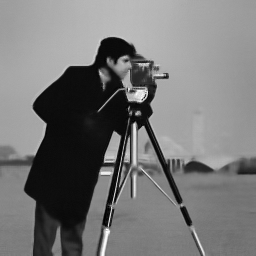} \\
(a) Clean image &&(b) Noisy image ($\sigma$= 20)&& (c) NLM using $v_l^{zero}$\\
\includegraphics[width=4cm]{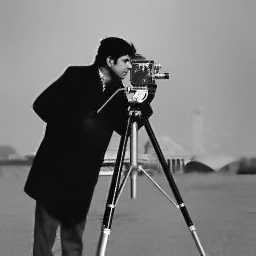} && \includegraphics[width=4cm]{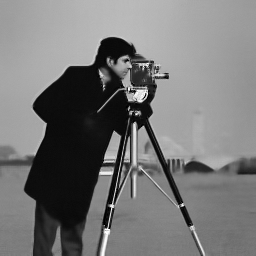} && \includegraphics[width=4cm]{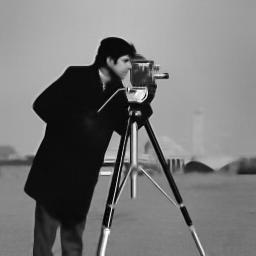} \\
(d) NLM using $v_l^{one}$ &&(e) NLM using $v_l^{stein}$&& (f) NLM using $v_l^{max}$\\
 \includegraphics[width=4cm]{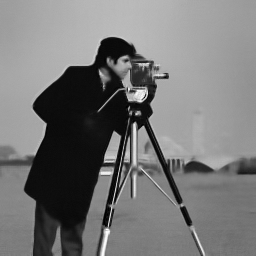}&& \includegraphics[width=4cm]{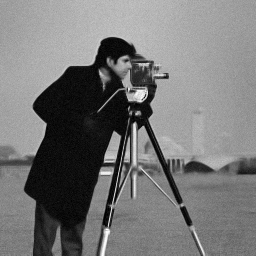} && \includegraphics[width=4cm]{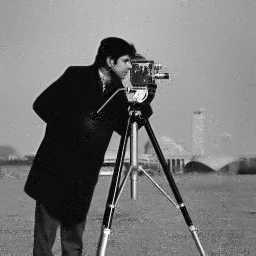} \\
(g) NLM using $v_l^{heur}$&&(h) NLM using $p^{*JS}$&& (i) NLM using $p_l^{*LJS}$\\
  \end{tabular}
  \caption{NLM denoising using various CPWs ($h\!=\!\sigma^2|\mathbb{P}|$, 7$\times$7 $\mathbb{P}$).}
\end{figure}
\section{Conclusion}
In this letter, we reviewed the CPW problem in NLM and proposed two new solutions JSCPW and LJSCPW based on the James-Stein estimator. We showed that the NLM-CPW problem can be viewed as the well studied statistical shrinkage estimator problem. This novel formulation opens a new door to the CPW problem and allows us to use the James-Stein shrinkage estimator for the NLM-CPW problem directly, that is the global JSCPW. To further enhance denoising performance, we propose a locally adapted James-Stein type CPW for each pixel. In this way, the denoising performance is tuned with respect to each local pixel rather than an entire image. Our experimental results show that the proposed James-Stein type CPWs help the NLM to achieve better overall performance in terms of a higher average PSNR with a more robust performance in terms of a smaller PSNR variance. By using these new CPWs, the NLM algorithm is then less sensitive to the temperature parameter $h$ and is better able to retain weak edges.

\bibliographystyle{IEEEtran}
\bibliography{report}
\end{document}